%% file: main.tex
\documentclass[preprint,5p,lefttitle]{elsarticle}

\usepackage{microtype}
\usepackage{amsmath}
\usepackage{amsfonts}
\usepackage{booktabs}
\usepackage{colortbl,multirow,multicol}
\usepackage{tabularx}
\usepackage{threeparttable}
\usepackage[edges]{forest}
\usepackage[hyperfootnotes=true,colorlinks,linkcolor=blue,citecolor=blue]{hyperref}
\usepackage{dblfloatfix}
\usepackage{caption}
\usepackage{makecell}
\usepackage{svg}
\usepackage{comment}
\usepackage{siunitx}
\usepackage{mhchem}
\newcommand{\hl}[1]{#1}
\usepackage{color}
\usepackage{cleveref}

\newcommand{\mathcolorbox}[2]{#2}

\begin{document}

\newcommand{\revisioncolor}{black}



\twocolumn[{
\renewcommand\twocolumn[1][]{#1}

\title{\textbf{Real-Time Data-Efficient Portrait Stylization via Geometric Alignment}}

\author[utokyo]{Xinrui Wang\corref{cor1}}
\ead{secret_wang@outlook.com}
\cortext[cor1]{Corresponding authors.}

\author[hat]{Zhuoru Li}
\author[meituan]{Xuanyu Yin}
\author[heshifan]{Xiao Zhou}
\author[utokyo]{Yusuke Iwasawa}
\author[utokyo]{Yutaka Matsuo}
\author[utokyo]{Jiaxian Guo}

\address[utokyo]{{The University of Tokyo, 7-3-1 Hongo, Bunkyo-ku}, Tokyo, 110-0033, Japan}
\address[hat]{Project Hat, Shenzhen, Guangdong province, China}
\address[meituan]{Meituan.co.ltd, Beijing, China}
\address[heshifan]{Hefei Normal University, Hefei, Anhui province, China

\begin{center}
    \captionsetup{type=figure}
    \includegraphics[width=0.95\textwidth]{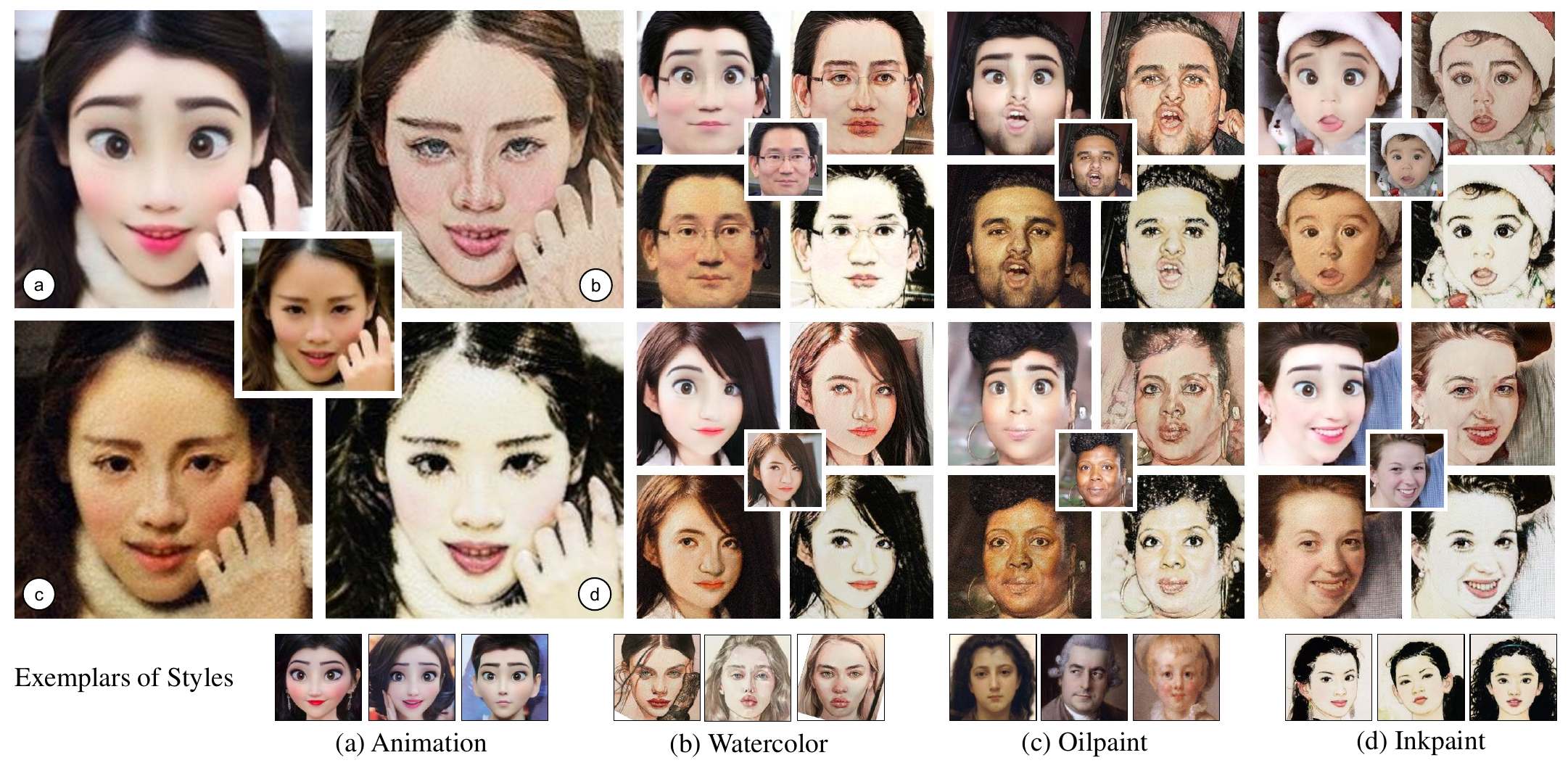}
    \vspace{-0.5em}
    \captionof{figure}{\small{Input portraits and stylized results of our proposed method. Leveraging geometric alignment, our approach can synthesize stylized portraits in real-time ($\sim30$ fps for $512\times512$ image) on mobile devices, even with limited style exemplars. Each example shows a portrait (center) and four stylized variants: Animation (a), Watercolor (b), Oilpaint (c), and Inkpaint (d).}}
    \vspace{-4em}
\end{center}
}

\input{0-abstract}

\maketitle

}]

\input{1-introduction}
\input{2-related-works}
\input{3-method}

\input{4-experiment}
\input{5-discussion}

 \vspace{-1em}
\bibliographystyle{elsarticle-num}
\bibliography{ref}  

\end{document}

%% file: 0-abstract.tex
\begin{abstract}

Portrait Stylization aims to imbue portrait photos with vivid artistic effects drawn from style examples. Despite the availability of enormous training datasets and large network weights, existing methods struggle to maintain geometric consistency and achieve satisfactory stylization effects due to the disparity in facial feature distributions between facial photographs and stylized images, limiting the application on rare styles and mobile devices.
 To alleviate this, we propose to establish meaningful geometric correlations between portraits and style samples to simplify the stylization by aligning corresponding facial characteristics. Specifically, we integrate differentiable Thin-Plate-Spline (TPS) modules into an end-to-end Generative Adversarial Network (GAN) framework to improve the training efficiency and promote the consistency of facial identities. By leveraging inherent structural information of faces, \emph{e.g.}, facial landmarks,  TPS module can establish geometric alignments between the two domains, at global and local scales, both in pixel and feature spaces, thereby overcoming the aforementioned challenges.
Quantitative and qualitative comparisons on a range of portrait stylization tasks demonstrate that our models not only outperforms existing models in terms of fidelity and stylistic consistency, but also achieves remarkable improvements in 2$\times$ training data efficiency and 100$\times$ less computational complexity, allowing our lightweight model to achieve real-time inference (30 FPS) at 512*512 resolution on mobile devices.

\end{abstract}


%% file: 1-introduction.tex
{\section{Introduction}\label{sec:introduction}}

\begin{figure*}[t]
\centering
\includegraphics[width=0.98\linewidth]{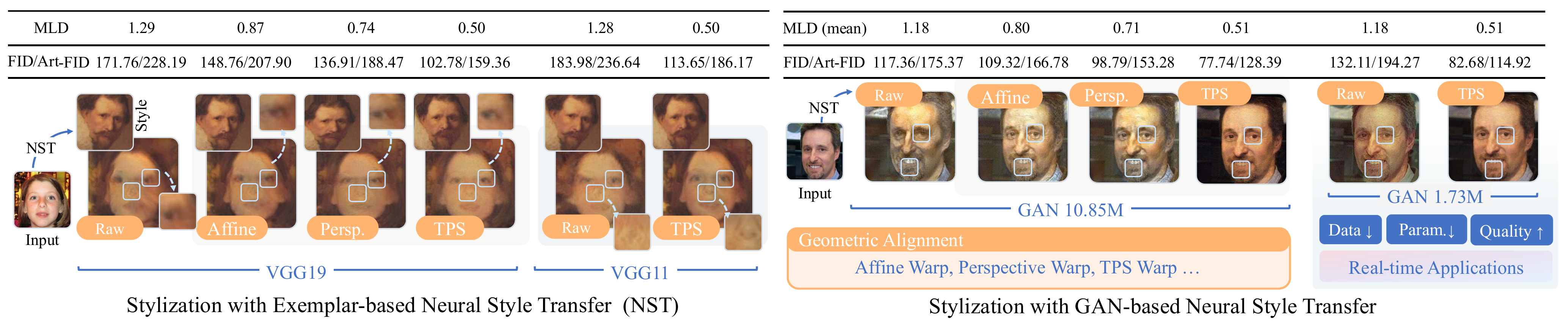}
\vspace{-0.5em}
\caption{\small{Effect of geometric alignment in stylization. We use Mean Landmark Distance (MLD) to measure geometric similarity between portrait-style image pairs and \hl{Frechet Inception Distance(FID)}/Art-FID to evaluate stylization quality. The stylization quality increases when portraits and style images become more geometrically similar. MLD, FID and Art-FID are described in Section \ref{section:experiment-setup}. Our experiments also show that geometric alignment enables smaller models with it to achieve comparable results with larger models without it. VGG11 (9M w/o head) and VGG19 (20M w/o head) are used as small and large model for exemplar-based NST and CycleGAN (10.85M) and its lightweight variant (1.73M) are used as small and large model for GAN-based style transfer.}}

\label{fig:alignment}
\vspace{-1.5em}
\end{figure*}

\begin{figure}[t]
\centering
\includegraphics[width=0.98\linewidth]{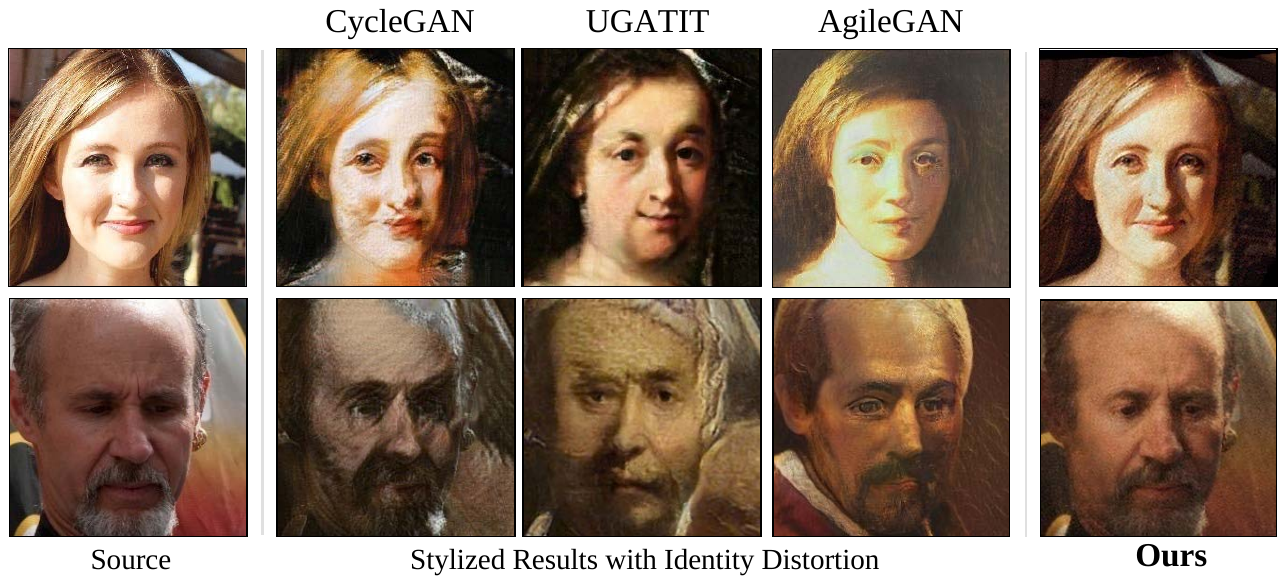}
\vspace{-0.5em}
\caption{\small{Illustration of identity distortion in portrait stylization tasks. All methods are trained on limited unpaired datasets.}}
 \label{fig:id_distortion}
\vspace{-1.5em}
\end{figure}

Portrait stylization casts vivid artistic visual effects to face photos from style examples \cite{shih2014style, selim2016painting}. It has long been a hot research topic \cite{shih2014style, selim2016painting, kim2019u, yi2020unpaired, texler2021faceblit, han2021exemplar}, the recent emerging of social networks and short video applications such as TikTok and Snapchat brings increasing popularity to it. Given the user preference for immediate responses and dynamic interactions on mobile devices, real-time capability is a highly desired feature. The demand for high-quality real-time portrait stylization methods has become a critical area of focus, garnering attention from both academic and industrial spheres.

Despite considerable advancements in previous portrait stylization methods\cite{shih2014style, song2021agilegan, yi2019apdrawinggan, yang2022pastiche} , a persistent challenge is the vast diversity in human appearance. Additionally, style samples often feature exaggerated artistic effects, leading to significant discrepancies in the shape and position of facial characteristics when compared to actual human portraits. The prevalent industry approach involves the manual creation of extensive paired datasets for pixel-level training, which is both labor-intensive and costly. Utilizing in-the-wild style datasets also poses challenges for GAN-based methods. Limited sample numbers often prove insufficient for the distribution matching process, while the entanglement of identity information and style information further compromises the stylization quality \cite{lee2020drit++} and causes identity distortion, which is shown in Fig.\ref{fig:id_distortion}.

We observed that portrait images share a consistent geometric structure represented by face landmarks, which are extensively utilized in facial image editing tasks. Based on this, we hypothesize that face landmarks can be leveraged to establish alignment between portrait and style image pairs, and expect it to enhance the stylization process, leading to improved synthesis quality and training efficiency, and reduced model capacity for faster inference. This hypothesis is tested with both exemplar-based and GAN-based neural style transfer (NST), and the results are shown in Fig.\ref{fig:alignment}.

The results Fig.\ref{fig:alignment} show that geometric alignment significantly enhances the stylization quality. In the order of original images to rigid affine transformation to perspective transformation to TPS warp, the geometric similarity evaluated by Mean Landmark Distance decreases by geometrically aligning the style images towards portraits, and the stylization quality measured by \hl{Frechet Inception Distance(FID)} and Art-FID increase for both exemplar-based and GAN-based NST. This indicates that improved geometric correspondence facilitates the stylization process and results in higher-quality stylized portraits.

Geometric alignment also accelerates inference. Recent literature notes the challenge of learning the correlation of geometry and textures between portraits and style images, often requiring computationally intensive methods such as attention mechanisms \cite{zhu2023all} and dense correspondence networks \cite{zhou2021cocosnet}. However, as is shown in Fig.\ref{fig:alignment}, Explicit geometric alignment with landmarks simplifies the stylization process between domains to reduce model capacity required, enabling smaller models with geometric alignment to outperform larger models without it, consequently accelerates inference. 

Moreover, geometric alignment enhances data efficiency. While large datasets are typically necessary for matching different distributions, geometric alignment reduces data needs by establishing spatial correspondence as a prior. During training, efficiency is also improved by randomly sampling and aligning portrait-style pairs of corresponding facial characteristics for local stylization. As identity information is primarily determined by facial geometry, the proposed geometric alignment restricts identity information during distribution matching by exclusively learning style information, thereby ensuring the consistency of identity in the resulting images and prevents identity distortion. This constraint further reduces style samples needed for distribution matching and improves data efficiency.

We further argue that the choice of geometric alignment techniques critically affects the stylization quality and computational demands.   Alignment in pixel-space \cite{shih2014style, chang2018pairedcyclegan, texler2021faceblit} may struggle with certain styles and are often limited to exemplar-based approaches. Our method, however, utilizes a flexible, differentiable TPS module for alignment in both image and feature spaces. This feature-space alignment, as shown in \cite{zhao2022thin}, enhances network representation and synthesis quality. We also demonstrate that feature-space deformation enables stylization of styles with large spatial deformation such as caricatures.

By integrating the differentiable geometric alignment module into an end-to-end GAN framework, our proposed method successfully bridges the distribution gap between portraits and style images and fulfills high-quality portrait stylization with light-weight networks and limited style datasets. To conclude, our contributions are as follows:

\begin{itemize}
\setlength{\itemsep}{0.5pt}
\setlength{\parsep}{0pt}
\setlength{\parskip}{0pt}

\item {In order to achieve fast and data-efficient portrait stylization, we hypothesized and proved that geometric alignment effectively bridges the gap between portraits and style images to enhance the stylization quality.}

\item {We designed a novel GAN framework incorporating a differentiable geometric alignment module, which facilitates end-to-end portrait stylization. It achieves satisfactory quality trained with less than 100 samples and inferences at realtime on mobile devices.}

\item {Qualitative comparison, quantitative evaluation and user study are conducted to show that our method outperforms previous works. Ablation study demonstrates the effectiveness of each component.}
\end{itemize}

%% file: 2-related-works.tex
\section{Related Works}
\subsection{Non-photorealistic Rendering}

Non-photorealistic Rendering (NPR)  brings artistic styles to photos. Iterative optimization based methods are used in early stage to imitate the painting textures of human artist, such as pencil drawing and sketches  \cite{lu2012combining, xu2011image}, oil-paint like brush stroke rendering \cite{hertzmann1998painterly}, and example based 2D style rendering \cite{wang2004efficient}.  CNN is later introduced for this task \cite{gatys2015neural, johnson2016perceptual}. I2I translation \cite{johnson2016perceptual, isola2017image, CycleGAN2017} are also widely adopted for NPR tasks, where people train transformation network for stylization, enabling inter-domain any to any transformation. Disentangle based methods \cite{lee2020drit++, huang2018multimodal} further try to separate style and content information in the images for better control during stylization.  What's more, techniques such as visual attention \cite{zhu2023all} and knowledge distillation \cite{zhu2020knowledge} are also proved helpful for NPR tasks.

While stylization may cause distortion and artifacts, some works attempted to employ spatial alignment to improve stylization quality, such as adopting a spatial relation augmented module into VGG network \cite{chang2021exploiting}, or performing rigid-alignment in feature space by treating features as point clouds in the channel dimension \cite{hada2021style}. But these methods only utilize spatial information implicitly with unsupervised attention mechanism, and the lack of semantic information as supervision also limits their performance.

In this study, we leverage the semantic information inherent in highly structured human faces to enhance the stylization process. Our method explicitly applies geometric alignment to both feature maps and images through the use of differentiable TPS transformations. This enables us to achieve geometrically aligned portrait stylization within a unified, end-to-end training framework.

\begin{table}[t]
\setlength\tabcolsep{5pt}
\begin{small}
\centering
\caption{\small{Comparison of existing portrait stylization methods. Numbers in scientific notation represent magnitude range. For prior methods, dataset sizes and inference times (at $512\times512$ resolution) are from original papers or official implementations. Prior deep learning methods require large datasets and extensive computation. Exemplar-based methods suffer from limited style diversity. Our approach, leveraging geometric alignment, performs comparably to exemplar-based methods while maintaining the style advantage of deep learning methods.}}
\begin{tabular}{ccccc}
\hline
\thead{Methods}&\thead{use \\ alignment}&\thead{alignment \\ space}&\thead{dataset \\ size}&\thead{inference \\ time(ms)}\\
\hline  
\thead{I2I translation \\ \cite{CycleGAN2017, lee2020drit++, kim2019u}}&\thead{no}&\thead{n/a}&\thead{$10^2\sim10^4$}&\thead{$\sim10^2$}\\
\hline 
\thead{StyleGAN \\ based \cite{song2021agilegan, yang2022pastiche}}&\thead{implicit}&\thead{latent \\ space}&\thead{$10^2\sim10^4$}&\thead{$10^2 \sim10^3$}\\
\hline
\thead{LDM+LoRA \\ \cite{zhang2023adding, rombach2022high}}&\thead{no}&\thead{n/a}&\thead{$10^1\sim10^3$}&\thead{$\sim10^3$ }\\
\hline
\thead{Few-shot I2I\\translation\cite{liu2019few}}&\thead{no}&\thead{n/a}&\thead{$10^0\sim10^1$}&\thead{$\sim10^2$}\\
\hline
\thead{Exemplar \\ based\cite{texler2021faceblit, shih2014style}}&\thead{explicit}&\thead{pixel \\ space}&\thead{1}&\thead{$10^1 \sim 10^3$}\\
\hline 
\thead{Makeup \\ transfer\cite{deng2021spatially, jiang2020psgan}}&\thead{explicit}&\thead{feature \\space}&\thead{$10^2\sim10^3$}&\thead{$\sim10^2$}\\
\hline 
\thead{\footnotesize{Ours}}&\thead{explicit}&\thead{pixel\& \\feature}&\thead{$10^1\sim10^2$}&\thead{$\sim10^1$}\\
\hline
\end{tabular}
\label{table:technics}
\end{small}
\vspace{-1em}
\end{table}

\vspace{-0.5em}
\subsection{Generative Models}

Generative Adversarial Network(GAN) \cite{goodfellow2014generative} synthesizes samples with the same distribution of training dataset by solving a min-max problem between a generator network and a discriminator network. It is powerful in image synthesis by forcing the generator and discriminator to reach Nash equilibrium with adversarial training and train converges when the generated images are indistinguishable from real images. Since proposed in 2014, it has been applied in various image synthesis tasks, including supervised image-to-image translation \cite{isola2017image}, unsupervised image-to-image translation \cite{CycleGAN2017} and can even disentangle the content and style information of the images\cite{huang2018multimodal}. Especially, StyleGANs \cite{karras2019style, karras2020analyzing} adopt a mapping network to map latent vector from normal distribution Z space to high-dimensional W space, and use it as style code to inject to convolution layers and control the synthesized results. They are not only capable of generating high quality samples, but also applicable to various downstream tasks \cite{pathak2016context, sanakoyeu2018style, zhang2018two, wang2020learning, zhang2021user}.  

Diffusion models \cite{song2019generative, ho2020denoising} recently achieved state-of-the-art performance on both sample quality \cite{dhariwal2021diffusion} and density estimation \cite{kingma2021variational} by physically modeling the diffusion process with a markov chain. Especially, Latent Diffusion Model \cite{rombach2022high} realized diffusion process in the low-resolution latent space, reduced computational complexity, increased synthesis quality and generalization. Moreover, DreamBooth \cite{ruiz2023dreambooth} and Low-Rank adaptation training (LoRA) \cite{hu2021lora} enables fine-tuning with small dataset and low computational cost, and ControlNet \cite{zhang2023adding} makes pixel aligned image-to-image translation possible for diffusion models. These techniques all expand the applicability of diffusion models and adopt them for diversified applications \cite{xu2024medsyn, xu2024ufogen}.

However, the parameter sizes and computation complexities for diffusion models are still too high to be applied in real-time scenarios. To enable real-time inference on mobile devices, we adopt adversarial leaning paradigm in an image-to-image (I2I) translation framework to train a light-weighted generator for portrait stylization.

\begin{figure*}[t]
\centering
\includegraphics[width=\linewidth]{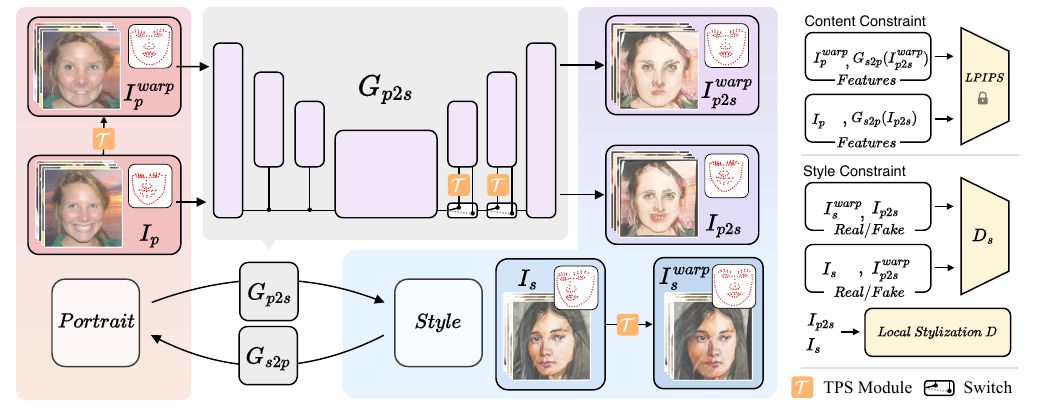}
\vspace{-2em}
\caption{\small{Overview of our framework. The proposed cycle-consistency framework involves two transformation directions: portrait to style and vice versa. For the portrait to style transformation, it comprises two branches: (1) the geometric warping branch, where the generator $G_{p2s}$ warps features from the portrait image $I_p$ using facial landmarks of the style image, synthesizing the aligned stylized result $I^{warp}_{p2s}$; and (2) the geometric invariant branch, where $G_{p2s}$ directly synthesizes the unaligned result $I_{p2s}$. The style image $I_s$ is warped with landmarks of the portrait to obtain $I^{warp}_s$. To constrain stylization, ($I^{warp}_{p2s}$, $I_s$) and ($I_{p2s}$, $I^{warp}_s$) are fed into the discriminator $D_s$, and $I_{p2s}$ is fed into the local stylization discriminator module. The cycle-consistency loss with frozen LPIPS is adopted to constrain content.}}
\vspace{-1em}
\label{fig:framework}
\end{figure*}

\vspace{-0.5em}
\subsection{Spatial Transformation in Image Synthesis}

WarpGAN\cite{shi2019warpgan} combined texture and geometric transformations to turn human faces into caricatures. However, the deformation was applied directly on pixel spaces, leading to artifacts and less precise results.
Deformable Style Transfer \cite{kim2020deformable} jointly optimizes the stylization and spatial warping process to achieve spatial transformation in style transfer, but the time-consuming optimization process limited its application. CocosNet \cite{zhou2021cocosnet} proposed a multi-stage patch-match in the feature space and Conv-GRU blocks to promote patch searching and matching performance. However, the synthesized portrait lost the original identity information, and the computational cost is also too high to be deployed on mobile devices. Guo \textit{et al.} \cite{guo2022alleviating} combined structure consistency constraint with cycle-consistent framework to improve I2I translation performance on similar domain images, but the implicit constraint still causes artifacts and is still sub-optimal for fine-grain control.

Image animation also benefits from spatial transformation. First order motion model \cite{siarohin2019first} at first learns a key points detector to establish spatial correspondence with jacobian matrix, the dense motion network then generates transformation map and occlusion map based on the detected spatial information and the decoder finally synthesizes the animated results with all information extracted. TPS motion model \cite{zhao2022thin} further improves and simplifies the image animation framework with differentiable TPS module, which directly establishes spatial deformation with only key points correspondence and directly applies to the feature maps. 

In this paper, we propose to use detected face landmarks to guide the differentiable TPS models to geometrically warp the portrait-style image pairs and enable the stylization to be learned in a spatial-aligned manner, this facilitates the style information to be transferred to the corresponding regions, importantly reducing the amount of style samples and number of parameters needed and improving the training efficiency.

\vspace{-0.5em}
\subsection{Portrait Stylization}

Portrait Stylization is a long standing topic in computer vision. Exemplar based methods \cite{shih2014style, texler2021faceblit, han2021exemplar} follow an align-transfer procedure, firstly spatially align the style example and then match the local statistics to the portrait. But these methods only perform one-to-one stylization and are not capable of learning the style information from a dataset, limiting their potential application. StyleGAN based methods \cite{yang2022pastiche, men2022unpaired, song2021agilegan} exploit the powerful synthesis ability of StyleGAN \cite{karras2019style}, but the result quality heavily relies on the precision of StyleGAN inversion, and also cause messy background. Makeup transfer is also well studied topic, \cite{jiang2020psgan} utilizes attention map and \cite{deng2021spatially} exploit segmentation mask to extract spatial information, both adopt additional style encoder to extract style code for latent space injection, which are too heavy for real-time applications. \cite{chang2018pairedcyclegan} exploit geometric warping for style samples, but it's only applied on the pixel space to help the discriminator training. 
In our method, we incorporate differentiable geometric alignment with the generator network, which performs explicit geometric alignment with facial landmarks and enables the model to be trained end-to-end and fulfill I2I translation between two distributions. The highly structured facial pattern of portrait images further encourage us to align and crop each facial feature for local stylization, enabling the model to be trained with limited style samples. Overall, we compare our method with existing methods in Table.\ref{table:technics}.

%% file: 3-method.tex
\vspace{-2em}
\section{Method}
We illustrate the overview of proposed framework in Fig.\ref{fig:framework}. The inputs of the framework are portrait images, style examples, and corresponding facial landmarks. We denote the portrait image, the style image, the portrait landmarks, and the style landmarks as $I_p$, $I_s$, $L_p$, $L_s$ respectively. We employ a 
There are two branches of forward pass during the generation stage: The geometric warping branch where the results are stylized by the generator network and also warped with TPS, and the geometric invariant branch where the results are only stylized by the generator network. The single direction portrait-to-style transformation is described as follows:

In the geometric warping branch, the generator $G_{p2s}$ warps the feature maps from $L_p$ to $L_s$ using integrated multi-scale TPS and synthesizes the deformed result $I^{warp}_{p2s}$, which has the identical landmark as $I_s$. In the geometric invariant branch, $G_{p2s}$ directly synthesizes geometrically unchanged $I_{p2s}$. TPS is applied to $I_s$ to warp it from $L_s$ to $L_p$ and get $I^{warp}_s$, which has identical landmark as $I_{p2s}$.

The two aligned image pairs are then fed into the discriminator $D_s$ to adversarially learn the mapping from the portrait domain to the style domain. A region-aware feature matching loss is adopted to force the synthesized results to match the statistics of style samples in aligned regions. We also randomly sample and crop style patches and align them to corresponding facial regions of synthesized results, and adopt auxiliary discriminators for local stylization.

The style-to-portrait transformation is strictly symmetric, and geometrically aligned cycle-consistency loss guarantees the translation cycle brings image back to the original domain. In the following sections, we introduce each component of the proposed framework in detail.

\vspace{-0.5em}
\subsection{Multi-scale TPS assisted generation}
To fulfill geometric deformation and edition, we integrate the TPS transformation \cite{bookstein1989principal} in the generator. TPS transformation is a nonlinear transformation that allows representing complex geometric deformation. Given corresponding landmarks of two images, we can warp one to the other with minimum distortion by applying TPS transformation $\mathcal{F}$:

\vspace{-1em}
\begin{equation}
  \label{eq:1}
  \begin{small}
    \begin{matrix}
     \begin{aligned}
     \min \iint_{R^{2}}\left(\left(\frac{\partial^{2} \mathcal{F}}{\partial x^{2}}\right)^{2}\right.+2\left(\frac{\partial^{2} \mathcal{F}}{\partial x \partial y}\right)^{2} +\left. \left(\frac{\partial^{2} \mathcal{F}}{\partial y^{2}}\right)^{2}\right) dx dy,
     \end{aligned}\\ 
   
    \text{s.t.} \quad \mathcal{F}(P^{\mathbf{P}}_i) = P^{\mathbf{S}}_i, \quad  i=1,2, \ldots, N,
  \end{matrix}
  \end{small}
\end{equation}

where $P^{\mathbf{P}}_i$ and $P^{\mathbf{S}}_i$ represent the landmarks of the portrait images and the style images respectively. For each source-target image pair, there are $N$ landmarks detected by predefined facial landmark detector (we set $N=10$ for our method), and generate one TPS transformation from portrait images $\mathbf{P}$ to style images $\mathbf{S}$. According to the derivation of equation \ref{eq:1} , the TPS transformation is obtained as below:

\vspace{-0.5em}
\begin{equation}
\label{eq:2}
  \mathcal{F}(p)=A\begin{bmatrix}
    p\\ 
    1   
    \end{bmatrix}+\sum_{i=1}^{N} w_{i} U\left({\left\lVert P^{\mathbf{S}}_{i}-p\right\rVert}_2\right),
\end{equation}

where $p=(x, y)^\top$ represents pixel coordinates, $A \in \mathcal{R}^{2\times 3}$ and $w_{i} \in \mathcal{R}^{2\times 1}$ are the TPS coefficients obtained by solving \cref{eq:1}, $U(r)$ is the radial basis function representing the influence of each landmark on the pixel at $p$:

\vspace{-0.5em}
\begin{equation}
\label{eq:3}
  U(r)=r^{2} \log r^{2}.
\end{equation}
\vspace{-1em}

\hl{We adopt an U-Net \mbox{\cite{ronneberger2015u}} like generator for stylization. The skip connections in U-Net help accelerate the model convergence, and also enhance the expressiveness for light-weight architecture with limited parameters.} In the geometric warping branch, we integrate TPS modules in every down-sampled scales to warp the feature maps. Each warped feature map is upsampled and concatenated to the next scale, until the 3 channels RGB images are synthesized. In the geometric invariant branch, the TPS modules are skipped and the stylization results are directly synthesized by the generator. When doing inference, the stylization results are synthesized directly by the generator without TPS, and TPS is used only when geometric editing is desired.

\subsection{Spatial aware discrimination}

The Adversarial loss of vanilla CycleGAN imposes constraint on the high-level semantic space, making it less controllable and easily affected by geometric distortion. In order to realize stylization and meanwhile preserve the identity and geometric appearance, we align $I_{p2s}$ and $I^{warp}_s$, $I^{warp}_{p2s}$ and $I_s$ as geometric identical portrait-style image pairs, and feed them to the discriminator for geometric aware discrimination. The adversarial loss for the portrait to style transformation $\min_{G_{p2s}} \max_{D_s} \mathcal{L}_{\text{GAN}}(G_{p2s},D_s)$ is expressed as below:

\vspace{-0.5em}
\begin{equation}
\label{eq:4}
\begin{aligned}
    \mathcal{L}_{\text{GAN}}(G_{p2s}, & \ D_s) =  \mathbb{E}_{I_s \sim p_{\text{data}}(I_s)}[\log D_s(I_s)]\\ + & \ \mathbb{E}_{I_p \sim p_{\text{data}}(I_p)}[\log (1-D_s(G_{p2s}(I_p))]
\end{aligned}
\end{equation}
\vspace{-0.5em}

We also adopt an adversarial loss for the symmetrical transformation: $\min_{G_{s2p}} \max_{D_p} \mathcal{L}_{\text{GAN}}(G_{s2p},D_p)$.

Feature matching loss is a commonly adopted technique in GAN models to stabilize training and improve synthesis quality. While previous methods match the statistics on the channel dimensions and neglect spatial information, we propose to reduce the channel dimension and match the statistics of the spatial dimension on the intermediate layer of the discriminator. The proposed geometric aware feature matching loss is presented as below:

\vspace{-1em}
\begin{equation}
\label{eq:7}
\begin{aligned}
  \mathcal{L}_{FM}(G_{p2s}, D_s) = [{||D^i_s(I_s)-D^i_s(G_{p2s}(I_p))||}_1]
\end{aligned}
\end{equation}
\vspace{-1em}

where $i$ represent the $i_{th}$ feature map of the discriminator. As the eyes, nose and mouth regions represent most of the information in the human face, it can fully exploit the aligned spatial information to improve the synthesis quality. We also adopt symmetric loss on the style to portrait transformation. In the experiments, we illustrate the geometric-aware feature-matching loss outperforms the vanilla feature-matching loss.

\begin{figure}[t]
\centering
\includegraphics[width=\linewidth]{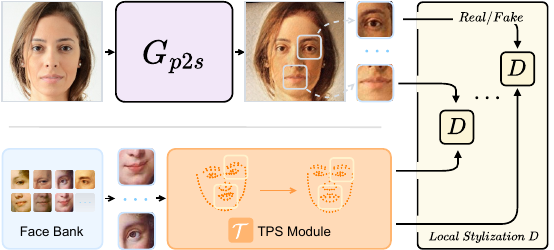}
\vspace{-1em}
\caption{Illustration of the local stylization D module. It utilizes a face bank containing cropped facial characteristics (eyes, nose, and mouth) from style images. During training, style patches are randomly sampled, geometrically aligned with the corresponding facial characteristics of portrait images, and fed into 4 auxiliary discriminators along with portrait patches. These discriminators enhance the stylization quality of facial characteristics.}
\vspace{-1em}
\label{fig:feature_bank}
\end{figure}

\subsection{Local alignment of facial characteristics for stylization}

Human face images are highly structured data in that everyone has the same facial characteristics. Inspired by the key observation that spatially aligning portrait-style image pairs effectively improve the stylization results, we further propose to crop and align corresponding facial characteristics, and employ auxiliary discriminators to learn local stylization. 

Shown in Fig.\ref{fig:feature_bank}, we crop the facial characteristics of style samples and save them with corresponding landmarks to a facial feature bank. During training, style patches of each face feature are randomly sampled and geometrically aligned to the corresponding facial characteristics of the result image. Four auxiliary discriminators are trained for left eye, right eye, nose and mouth regions to distinguish whether the patches are real or fake and facilitate the stylization quality. Similar to Eq.\ref{eq:4}, we formulate the auxiliary GAN loss as:

\vspace{-1em}
\begin{equation}
\label{eq:auxiliary}
\begin{aligned}
    \mathcal{L}_{\text{auxiliary}}(G, D_j) = & \ \sum_{D_j \in D}\mathbb{E}_{(I_s, I_p ) \sim p_{\text{data}}}  [[\log D_j(I_s)]\\ + & \ [\log (1-D_j(G(I_p))]]
\end{aligned}
\end{equation}
\vspace{-0.5em}

The design of randomly sampling style patches from the face feature bank to match content images for stylization greatly increases the combinations of content-style pairs, bringing extra diversity into training, and thus severely reduced the number of style samples needed.

\subsection{Full model}

We adopt a cycle-consistency framework to constrain the appearance change during the portrait stylization. To avoid the sub-optimal results caused by large geometric gaps between two domains, we only apply cycle-consistency loss on the geometric invariant branch, where the image triplets involved in calculating cycle-consistency loss have identical landmarks. Different from previous methods that directly minimize the L1 loss on images, we adopt Learned Perceptual Image Patch Similarity (LPIPS) loss \cite{zhang2018unreasonable} which was calculated with a image recognition network pretrained on perceptual patch similarity dataset and helps accelerate the convergence and improves synthesis quality. We show the cycle-consistency as below:

\vspace{-0.5em}
\begin{equation}
\label{eq:5}
\begin{aligned}
   \mathcal{L}_{\text{CYC}} =  & \ \mathbb{E}_{I_s\sim p_{\text{data}}(I_s)}||G_{p2s}(G_{s2p}(I_s))-I_s||_{lpips}
    \\ + & \ \mathbb{E}_{I_p\sim p_{\text{data}}(I_p)}||G_{s2p}(G_{p2s}(I_p))-I_p||_{lpips}
\end{aligned}
\end{equation}
\vspace{-0.5em}

The full model is trained by jointly minimizing the following losses:
\begin{equation}
\label{eq:8}
\begin{small}
\begin{aligned}
\mathcal{L}(G_{p2s}, G_{s2p}, D_s, & D_p) = \mathcal{L}_{GAN}(G_{p2s}, G_{s2p}, D_s, D_p)\\
								&+ \lambda_1* \mathcal{L}_{\text{auxiliary}}(G, D_j) + \lambda_2* \mathcal{L}_{\text{CYC}}\\
								&+ \lambda_3* \mathcal{L}_{FM}(G_{p2s}, G_{s2p}, D_s, D_p)\\ 				
\end{aligned}
\end{small}
\end{equation}
Where $\lambda_1$,  $\lambda_2$ and  $\lambda_3$ represent the weight of each term.

\begin{figure}[t]
\centering
\includegraphics[width=0.95\linewidth]{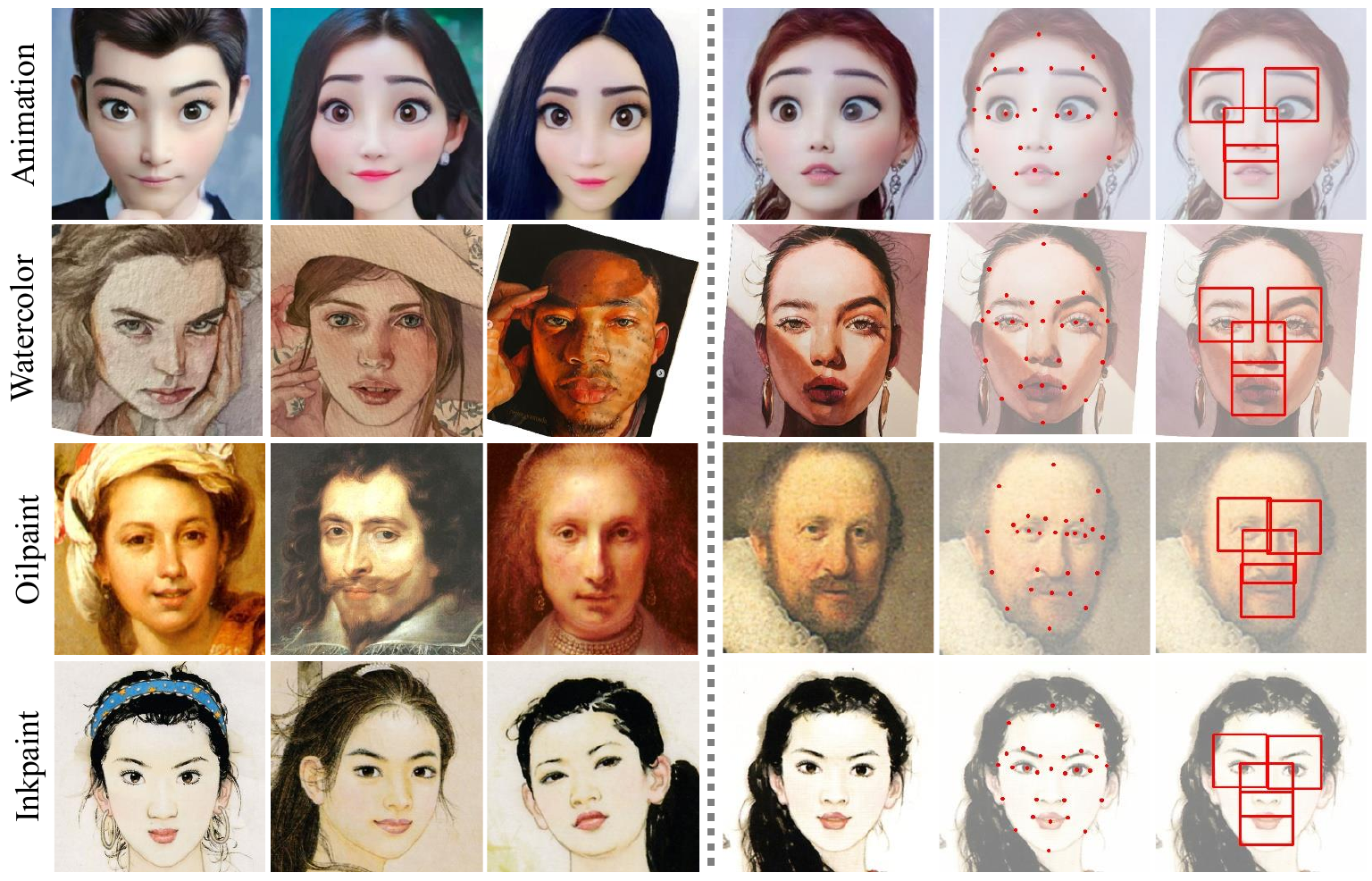}
\vspace{-1em}
\caption{Overview of our dataset. We show examples from 4 different styles, from top to bottom: Animation (N=800), Watercolor (N=64), Oilpaint (N=300) and Inkpaint (N=34). \hl{In the right part, we illustrate the landmarks used for TPS warping and the facial characteristics regions cropped out for local stylization.}}
\vspace{-1em}
\label{fig:dataset}
\end{figure}

%% file: 4-experiment.tex
\begin{figure*}[t]
\centering
\includegraphics[width=\linewidth]{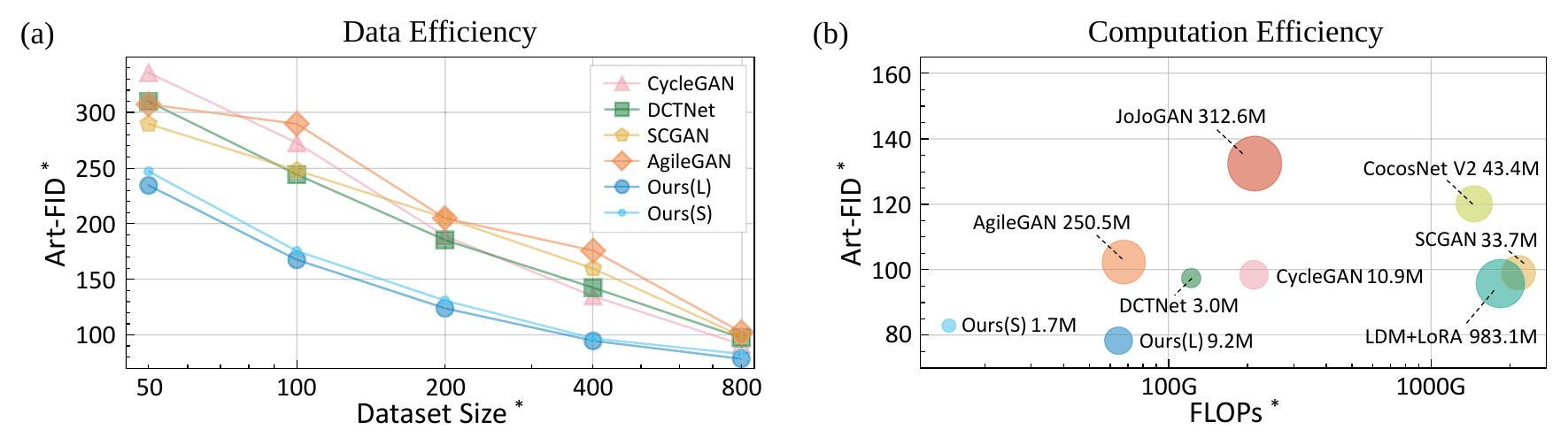}
\vspace{-1.5em}
\caption{Comparisons on data efficiency and computation efficiency. We present the evolution of Art-FID on Animation dataset when scaling dataset size (a) and Art-FID versus FLOPs (b) of different methods. Lower Art-FID represents better performance. Circle size (b) represents the model parameters scale. * indicates that smaller values are preferable. Our method achieves $\sim2\times$ better data efficiency and $\sim100\times$ less computational complexity compared to existing approaches.}
\label{fig:line}
\vspace{-1.5em}
\end{figure*}

\section{Experiment}

\subsection{Experiment Setup}
\label{section:experiment-setup}

\textbf{Implementation.} The proposed method is implemented by Pytorch \cite{paszke2019pytorch}. We describe the implementation details in the supplementary material. Adaptive discriminator augmentation \cite{karras2020training} is adopted to prevent overfitting. The learning rate and batch size are set to be $1\times10^{-4}$  and 1. Adam optimizer \cite{kingma2014adam} is adopted to optimize both networks. Training stops at 300000 step or on convergency. 

\textbf{Dataset.} We train the stylization network on 4 different styles and 2 different resolutions. We cropped out the facial regions of style samples and resized them to $256\times256$ and $512\times512$ resolutions respectively. For the portrait data, we collect 10000 images from the FFHQ dataset \cite{karras2019style} and resize to different resolutions. For the \textbf{oil paint} style, we collect 300 facial images from the Wikiart dataset \cite{nichol2016painter}. For the \textbf{animation} style, the \textbf{water color} style and the \textbf{ink paint style}, we collect 800 images, 64 images and 34 images respectively. \hl{To obtain the facial landmarks, we employ an industry level facial landmark detector to extract 228 facial landmarks, and select 28 landmarks for TPS warping. We further crop the eyes, nose and mouth regions based on the landmarks as the input of  auxiliary discriminators for local stylization. The size of crop regions are $64\times64$ for $256\times256$ images and $128\times128$ for $512\times512$ images. The examples of dataset are illustrated in Figure \mbox{\ref{fig:dataset}}. All datasets together with landmarks and facial region coordinates will be made publicly available to facilitate further studies.}

\begin{table*}[t]
\setlength\tabcolsep{2.5pt}
\begin{small}
\centering
\caption{Quantitative comparison evaluated by FID/Art-FID. \textcolor{red}{Red} and \textcolor{blue}{Blue} represent the 1st and 2nd place performance, respectively. For LDM with LoRA, the parameters and FLOPs of the text encoder were not calculated. For DCT-Net, only the parameters and FLOPs of the head stylization network and background stylization network were calculated. For VCT, the FLOPs is not available as iterative inversion method is used.}
\begin{tabular}{ccccccccccc}
\hline
{Methods}&{Cyclegan}&{Agilegan}&{SCGAN}&{CocosNetV2}&{DCTNet}&{LDM}&{JoJoGAN}&{VCT}&{Ours(L)}&{Ours(S)}\\
\hline  
{Animation}&{\tiny{200.56/98.47}}&{\tiny{60.32/102.36}}&{\tiny{112.57/99.07}}&{\tiny{132.10/120.18}}&{\tiny{103.27/97.46}}&{\tiny{107.65/95.79}}&{\tiny{150.57/132.66}}&{\tiny{75.49/110.28}}&{\tiny{\textcolor{red}{40.48/78.36}}}&{\tiny{\textcolor{blue}{47.61/82.93}}}\\
\hline 
{Watercolor}&{\tiny{180.09/355.34}}&{\tiny{162.61/312.12}}&{\tiny{150.97/266.18}}&{\tiny{\textcolor{blue}{138.14}/291.78}}&{\tiny{144.79/155.87}}&{\tiny{161.29/261.16}}&{\tiny{231.11/259.20}}&{\tiny{187.09/271.78}}&{\tiny{\textcolor{red}{132.89/209.18}}}&{\tiny{141.77/\textcolor{blue}{217.30}}}\\
\hline
{Oilpaint}&{\tiny{117.36/175.37}}&{\tiny{101.71/204.65}}&{\tiny{116.94/187.56}}&{\tiny{119.80/216.23}}&{\tiny{101.26/165.29}}&{\tiny{108.36/177.23}}&{\tiny{156.07/235.51}}&{\tiny{121.76/185.98}}&{\tiny{\textcolor{red}{77.74/128.39}}}&{\tiny{\textcolor{blue}{82.68/144.92}}}\\
\hline
{Inkpaint}&{\tiny{150.75/327.64}}&{\tiny{146.08/351.28}}&{\tiny{126.93/292.37}}&{\tiny{156.26/302.71}}&{\tiny{139.80/290.18}}&{\tiny{168.34/315.48}}&{\tiny{175.73/299.12}}&{\tiny{147.76/292.08}}&{\tiny{\textcolor{red}{100.42/233.27}}}&{\tiny{\textcolor{blue}{105.28/249.65}}}\\
\hline
{Param (M)}&{10.85}&{250.50}&{33.69}&{43.39}&{\textcolor{blue}{2.95}}&{983.07}&{312.60}&{983.07}&{9.15}&{\textcolor{red}{1.73}}\\
\hline
{FLOPs (G)}&{211.84}&{67.63}&{2149.58}&{1462.17}&{122.02}&{1837.12}&{215.65}&{N/A}&{\textcolor{blue}{64.59}}&{\textcolor{red}{14.60}}\\
\hline
\end{tabular}
\vspace{-1em}
\label{table:fid}
\end{small}
\end{table*}

\textbf{Evaluation Metrics.} In qualitative comparison, we show the results of our proposed method and previous methods with qualitative analysis. In quantitative evaluation, 
\hl{we use both FID \mbox{\cite{heusel2017gans}} and \mbox{Art-FID \cite{wright2022artfid}} as the criterion for quantitative evaluation. FID is a criterion designed to quantitatively evaluate the performance of generative models by measuring the distance of distribution.  An image classification network \mbox{\cite{szegedy2016rethinking}} is trained on a large-scale artwork dataset and used to extract high-level features of images to calculate the distance between two distributions. Consequently, lower FID scores and Art-FID scores represent  better performance. We denote content images as $X_c$, style images as $X_s$ and stylized images as $X_g$, FID calculation is formulated as:}

\vspace{-2em}
\begin{equation}\label{eq:fid}
\begin{small}
    \mathcolorbox{yellow}{
    \textrm{FID}(X_s, X_g) = ||\mu_s - \mu_g||_2^2 + \textrm{Tr}(\Sigma_s + \Sigma_g - 2 (\Sigma_s \Sigma_g)^\frac{1}{2}),
    }
\end{small}
\end{equation}
\vspace{-1em}

\hl{Where $(\mu_s, \Sigma_s)$ and $(\mu_g, \Sigma_g)$ are the mean and covariance of the Inception features of the style images $X_s$ and the stylized images $X_g$, respectively. Art-FID is a modified version of FID of style transfer tasks, which takes content preservation as consideration, and can be formulated as:}

\vspace{-1em}
\begin{equation}\label{eq:art_fid}
    \mathcolorbox{yellow}{
   \textrm{ArtFID}(X_g, X_c, X_s) = \Bigg( 1 + \frac{1}{N} \sum\limits_{i=1}^N d(X_c^{(i)}, X_g^{(i)}) \Bigg) \cdot \Bigg( 1 + \textrm{FID}(X_s, X_g) \Bigg)
    }
\end{equation}
\vspace{-1em}

To evaluate the geometric deformation between two portrait images, we also define a quantitative metric called mean landmark distance (MLD), smaller MLD means two images are more similar:

\vspace{-1em}
\begin{equation}
  \label{eq:mle}
  \begin{small}
  \begin{aligned}
   \mathcal{F}_{\text{MLD}}(I_a, I_b) = \sqrt{\frac{\sum_{i=0}^N{[(x^a_i-x^b_i)^2 + (y^a_i-y^b_i)^2]}}{N*h*w}}
  \end{aligned}
  \end{small}
\end{equation}

It represents the normalized mean value of euclidean distances between the corresponding landmarks of two portrait images, where $I_a$ and $I_b$ represent two portrait images, N represents the total number of landmarks, h and w represent the height and width of the image, and $x^a_i$, $y^a_i$ represent the coordinates of a landmark. When calculating MLD in this paper, we resize the two images to the same size and extract 10 landmarks in corresponding positions, which are the same as the landmarks we used for training.

\begin{table}[t]
\setlength\tabcolsep{2pt}
\centering
\caption{Comparison of inference time. Our method shows the fastest inference time across different devices. The lightweight model Ours(S) achieves real-time inference at 33 FPS on a mobile device. Note that only a subset of previous methods are compared due to the lack of support for certain operators in open-source mobile inference engines.}
\begin{small}
\begin{tabular}{ccccccc}
\hline
{}&\makecell{\tiny{CycleGAN}}&\makecell{\tiny{DRIT++}}&\makecell{\tiny{UGATIT}}&\makecell{\tiny{SCGAN}}&\makecell{\tiny{Ours(L)}}&\makecell{\tiny{Ours(S)}}\\
\hline  
{\tiny{Ryzen6900(ms)}}&{572}&{697}&{716}&{1423}&{219}&{62.5}\\
\hline 
{\tiny{RTX4070m(ms)}}&{154}&{188}&{185}&{354}&{53.1}&{12.4}\\
\hline
{\tiny{Snapdragon 8Gen1(ms)}}&{319}&{366}&{378}&{633}&{114}&{31.8}\\
\hline
\end{tabular}
\label{table:speed}
\end{small}
\end{table}

\textbf{Time Performance.} We train both a large generator with ~9M parameters and a small generator with ~1.7M parameters using the proposed framework, and evaluate the inference speed on three devices: AMD Ryzen 6900 laptop CPU, Nvidia RTX4070 laptop GPU and Qualcomm Snapdragon 8Gen1 mobile SOC. We show the comparison of inference speed on different devices at $512\times512$ resolution in table \ref{table:speed}. Only part of previous methods are compared as the rest of them use operators not supported by open-source inference engine for mobile devices. From \ref{table:speed} we can clearly see that the inference speed of proposed small model is one order-of-magnitude faster can previous methods and can infer at real-time on both GPU and mobile phone.

\begin{figure*}[t]
\centering
\includegraphics[width=\linewidth]{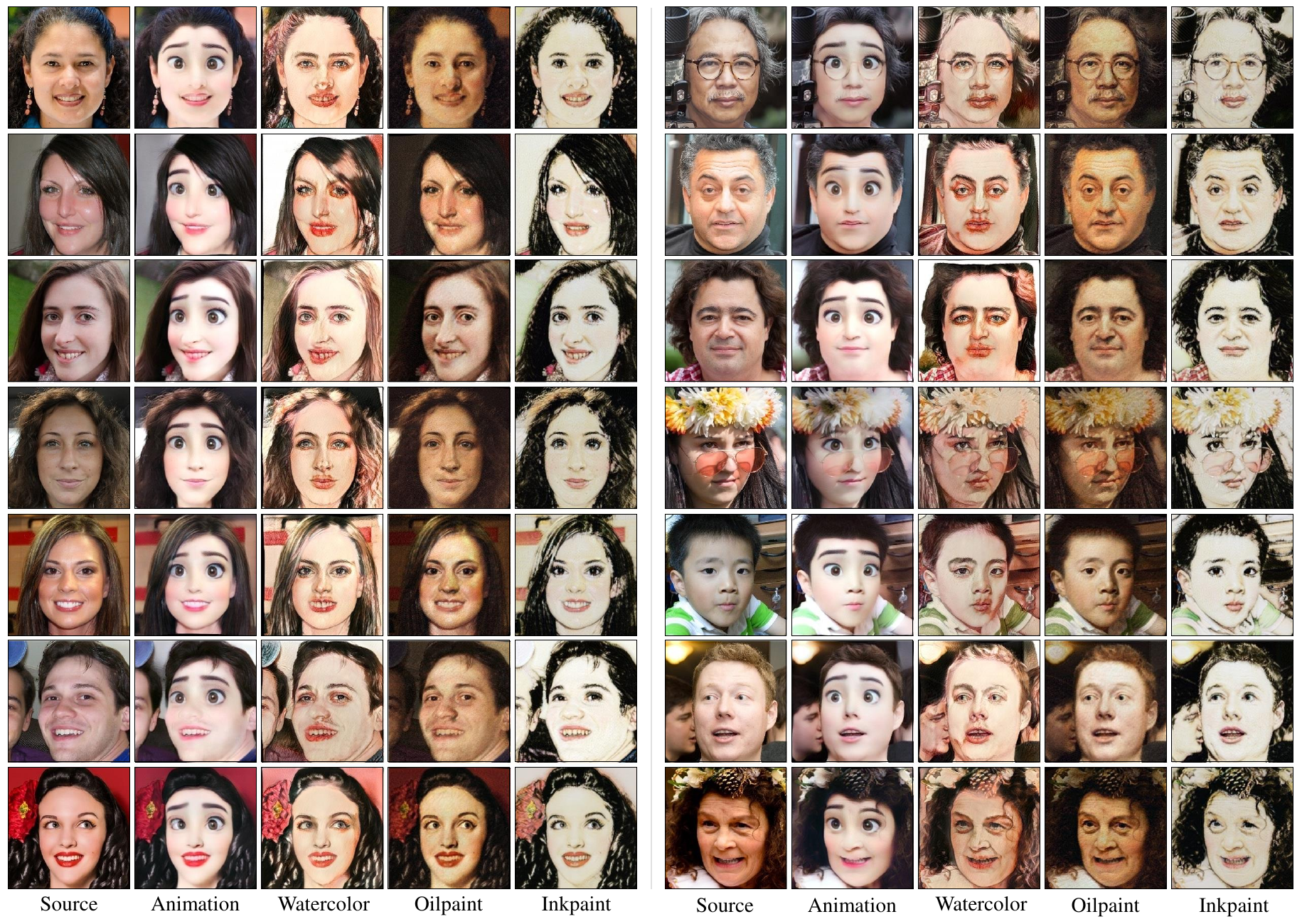}
\vspace{-2em}
\caption{Results of synthesized portraits of \textbf{our method} in various styles. Our method excels in preserving intricate content and styles, effectively handling complex portraits with occlusions, exaggerated expressions, and makeup.}
\vspace{-1em}
\label{fig:results}
\end{figure*}

\begin{figure*}[htb]
\centering
\includegraphics[width=\linewidth]{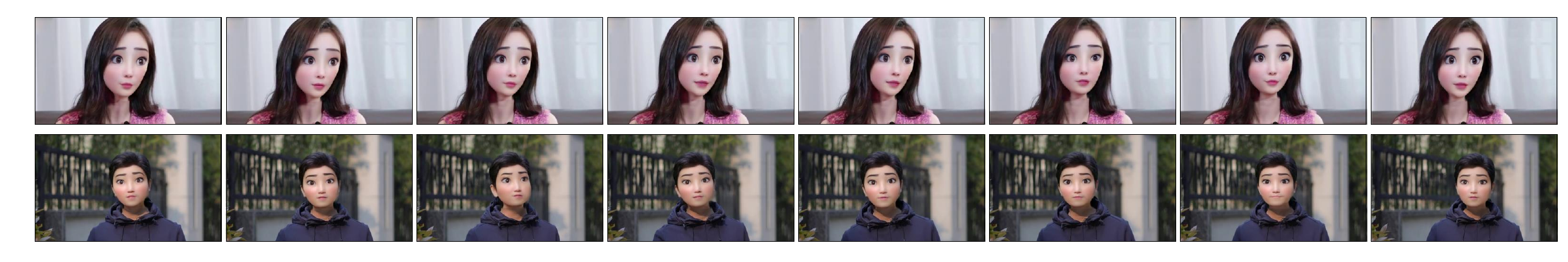}
\vspace{-2em}
\caption{Results of stylized video frames in Animation style. Our method synthesizes temporally coherent frames that preserve both content and artistic style.}
\vspace{-1em}
\label{fig:frames2}
\end{figure*}

\textbf{Illustration of Stylized Results.}
Four models are trained for 4 different style images in the dataset respectively. The stylization results of different models are shown in Fig.\ref{fig:results}. We also show two stylized video frame sequences in Fig.\ref{fig:frames2}, where the facial characteristics are lively stylized and the facial expressions are loyally preserved, and the stylized faces are smoothly merged into the background.

\begin{figure}[t]
\centering
\includegraphics[width=\linewidth]{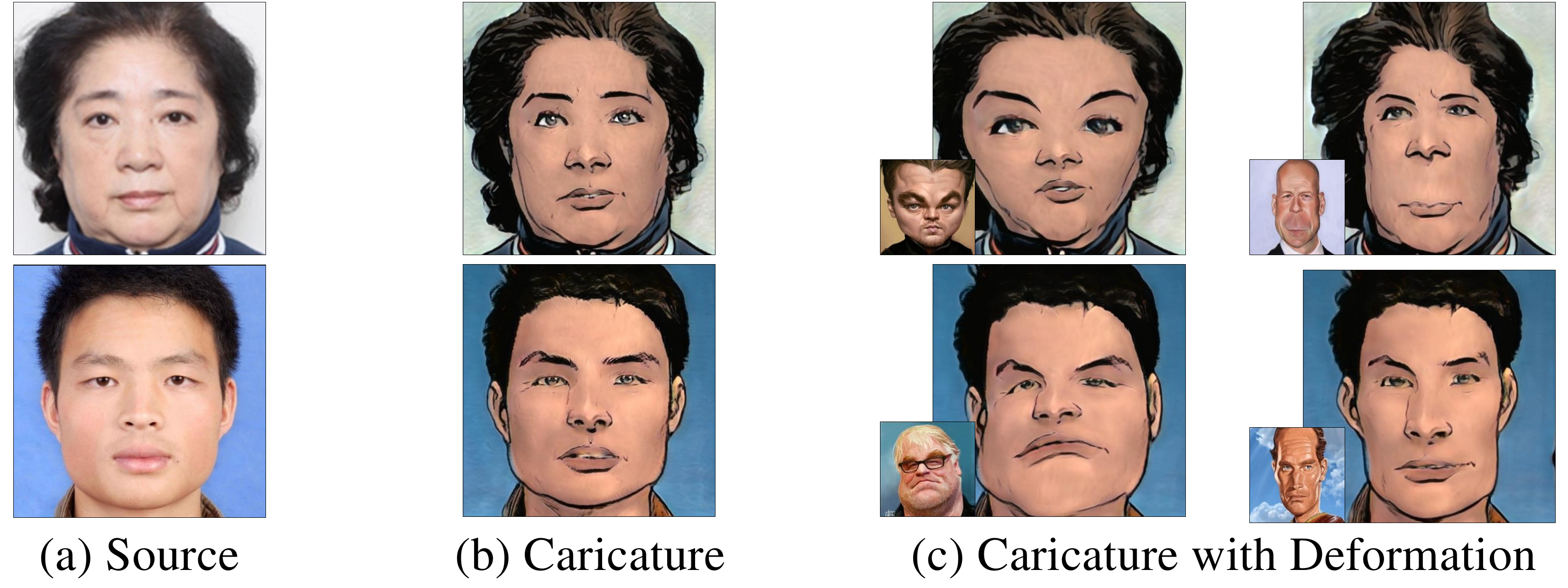}
\vspace{-2em}
\caption{Results of synthesized portraits in Caricature style. The bottom-left corner of each deformation result shows the geometric reference. As demonstrated by the results, our method can handle styles with large geometric deformations. }
\vspace{-1em}
\label{fig:caricature}
\end{figure}

\textbf{Stylization Results with Large Deformation.}
We trained a model for caricature style to illustrate the proposed method's stylization ability with large geometric deformation. A facial landmark detector for artistic style \cite{yaniv2019face} was deployed to detect the landmarks of caricature images. We illustrate the results in Fig.\ref{fig:caricature}, the portrait images are shown in column a, the direct stylized results are shown in column b, and geometrically deformed stylization results are shown in column c with corresponding caricature reference images in the bottom-left corner. The results are stylized with vivid artistic style and meanwhile maintain the original identity information, and can also be geometrically deformed with reference images. This proves that our proposed method is capable of stylizing portrait images with severely deformed non-photorealistic style examples.

\begin{figure*}[p]
\centering
\includegraphics[width=\linewidth]{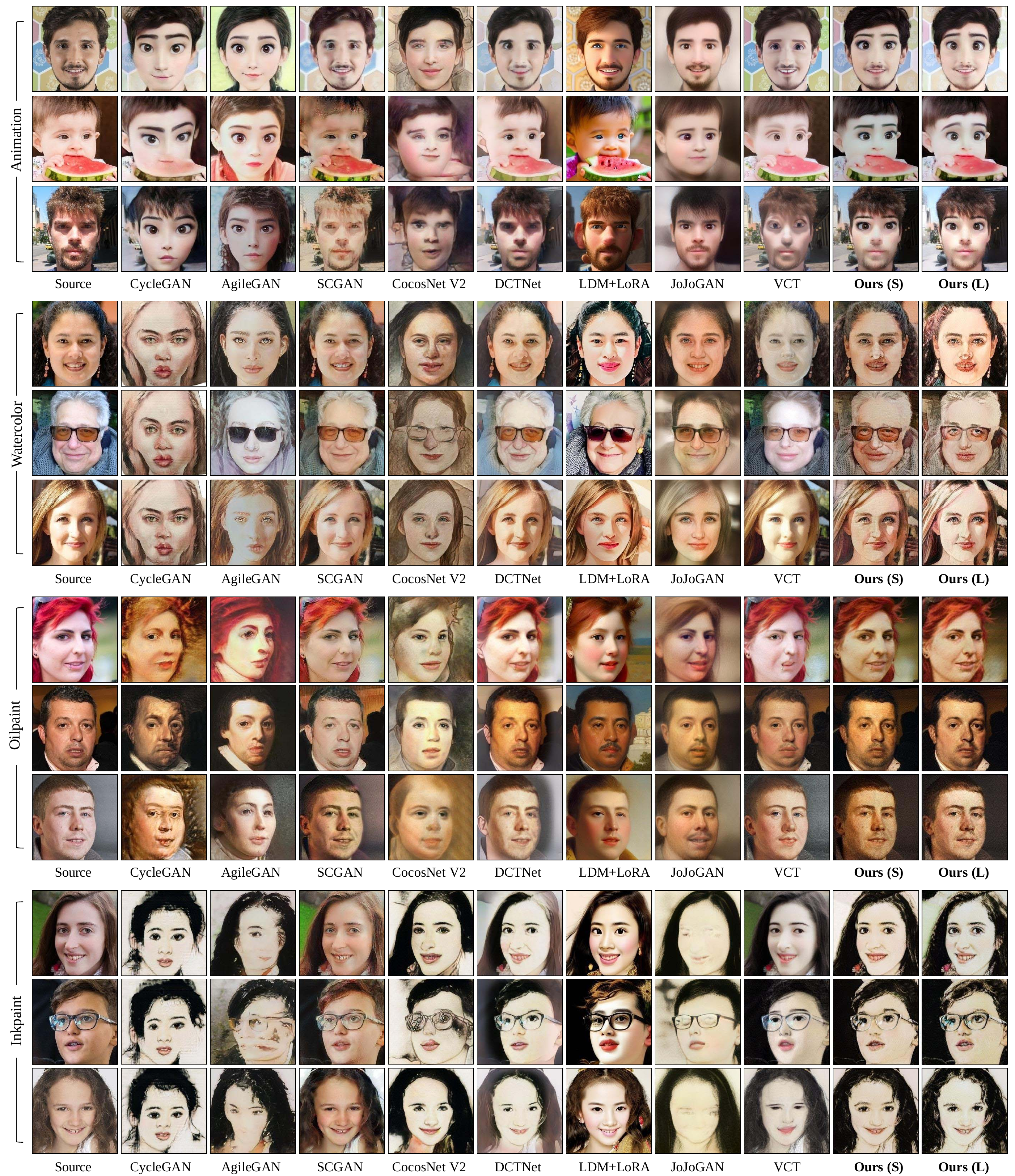}
 \vspace{-1em}
\caption{Qualitative comparison with existing methods. We use Ours(S) and Ours(L) to denote our lightweight model and full model respectively.}
\label{fig:comparison}
\end{figure*}

\textbf{Comparison.} For single image stylization, we compare our framework with CycleGAN \cite{CycleGAN2017} that represents unpaired I2I translation, AgileGAN\cite{song2021agilegan} and JoJoGAN\cite{chong2022jojogan} that represents StyleGAN based stylization, SCGAN \cite{deng2021spatially} that represents makeup transfer, DCTNet\cite{men2022dct} that represents portrait cartoonization, CocosNet\cite{zhou2021cocosnet} that represent geometric alignment, LDM (Latent Diffusion Model) \cite{rombach2022high} with Low-Rank adaptation training (LoRA) \cite{hu2021lora} and ControlNet \cite{zhang2023adding} that represents diffusion models, and VCT \cite{cheng2023general} that represent diffusion inversion in Fig. \ref{fig:comparison}.  All methods are trained with the official implementations, default settings, and the same dataset used for our method.

We also conducted experiments on DRIT++ \cite{lee2020drit++} for example based generation and U-GAT-IT \cite{kim2019u} for unpaired portrait stylization. The inference speeds are compared in Table \ref{table:speed}, but the qualitative and quantitative performance are not as good as the methods shown in Fig \ref{fig:comparison}. Due to the limitation of space, we show the results in supplementary materials. 

\begin{figure*}[htb]
\centering
\includegraphics[width=\linewidth]{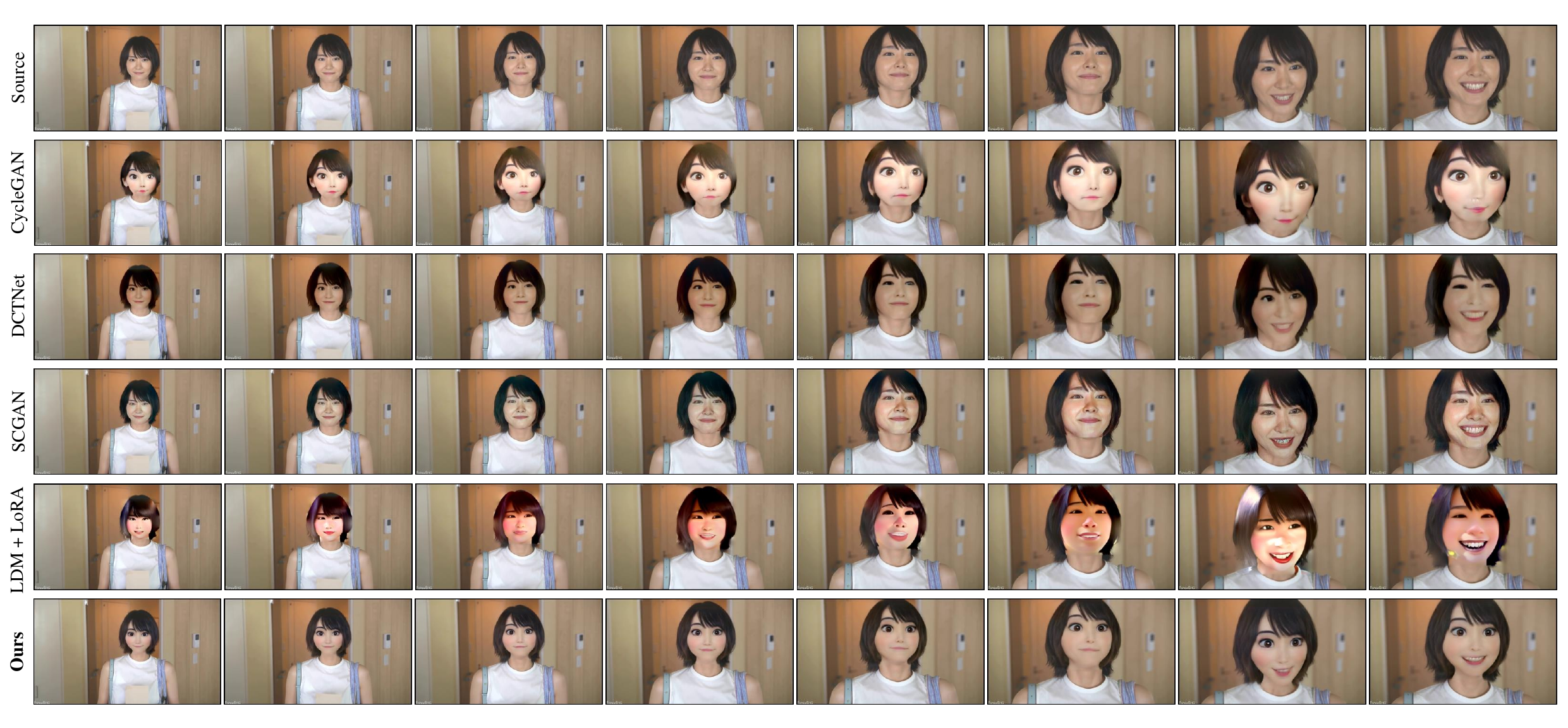}
\vspace{-2em}
\caption{Qualitative comparison of video frame sequence stylization in Animation style with state-of-the-art methods.}
\vspace{-1em}
\label{fig:frames1}
\end{figure*}

For video results, we show the comparison of stylized video frame sequences of CycleGAN, SCGAN, DCTNet and LDM with LoRA and our proposed method in Fig.\ref{fig:frames1}.

\vspace{-1em}
\subsection{Quantitative evaluation}


In this work,  FID \cite{heusel2017gans} and Art-FID \cite{wright2022artfid}  are adopted to evaluate the performance of previous methods and our method. As is shown in table \ref{table:fid}, our large model achieves 1st in both FID and Art-FID of all 4 styles, and our lightweight model achieves 2nd in Art-FID of 4 styles and 2nd in FID of 3 styles and 3rd of watercolor style, indicating that our proposed method synthesizes results similar to style examples and also achieves high quality.

Our detailed quantitative analysis, depicted in Fig.\ref{fig:line}, underscores the data and computational efficiency of our proposed method. On the left side of the figure, we compare performance of different methods across varying dataset sizes. Notably, both our large and small models are positioned at the base of the graph, indicating that our method outperforms others across all dataset sizes. Moreover, a cross-comparison reveals that our models, when trained on half the dataset size, either match or surpass the performance of other methods. This highlights the data efficiency inherent in our approach.

On the right side of Fig.\ref{fig:line}, we contrast performance in relation to model size and floating-point operations per second (FLOPs). Our large and small models achieve the lowest Art-FID scores while maintaining the smallest model size and FLOPs, a testament to their computational efficiency. The figure also indicates that previous methods, with performance close to ours, incur two orders of magnitude more computational costs. This stark difference underscores the computational advantages of our method, enabling it to inference at real-time on mobile devices.

\begin{figure*}[t]
\centering
\includegraphics[width=0.98\linewidth]{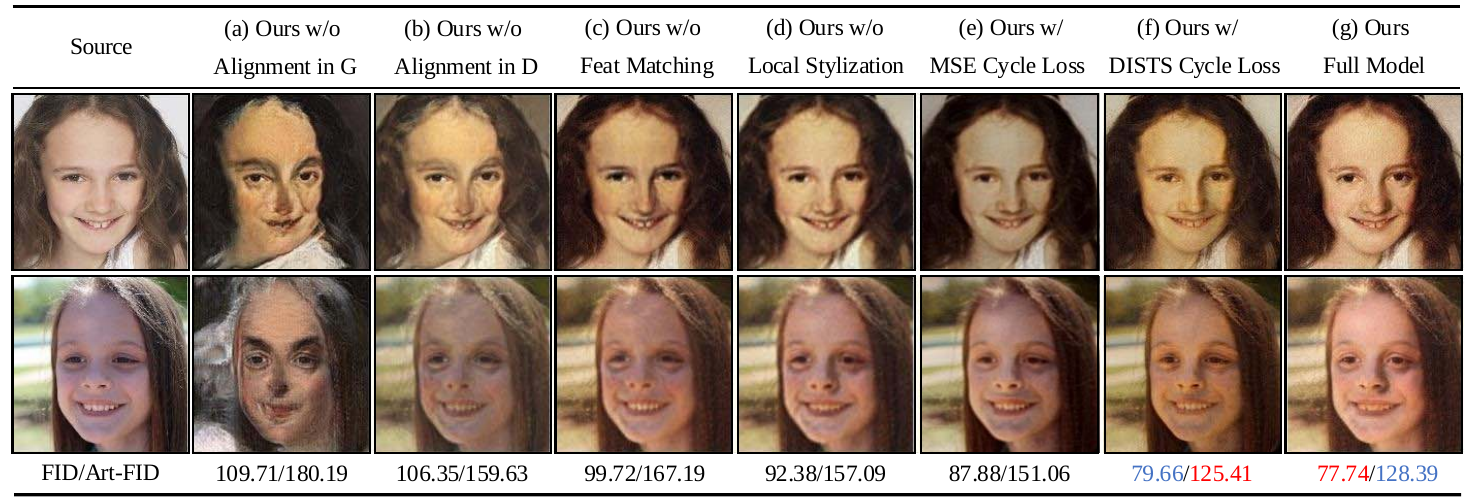}
\vspace{-1em}
\caption{Ablation study on individual components. (a) Ablating alignment in generators $G$; (b) Ablating alignment in discriminators $D$; (c) Ablating geometric-aware feature-matching loss; (d) Ablating local stylization D module; (e) Use MSE cycle loss; (f) Use DISTS cycle loss; (e) Our proposed full model. We also show quantitative results evaluated by FID/Art-FID \textcolor{red}{Red} and \textcolor{blue}{Blue} represent 1st and 2nd place in performance. The results are calculated on oilpaint dataset.}
\label{fig:component}
\vspace{-1em}
\end{figure*}

\begin{figure}[t]
\centering
\includegraphics[width=\linewidth]{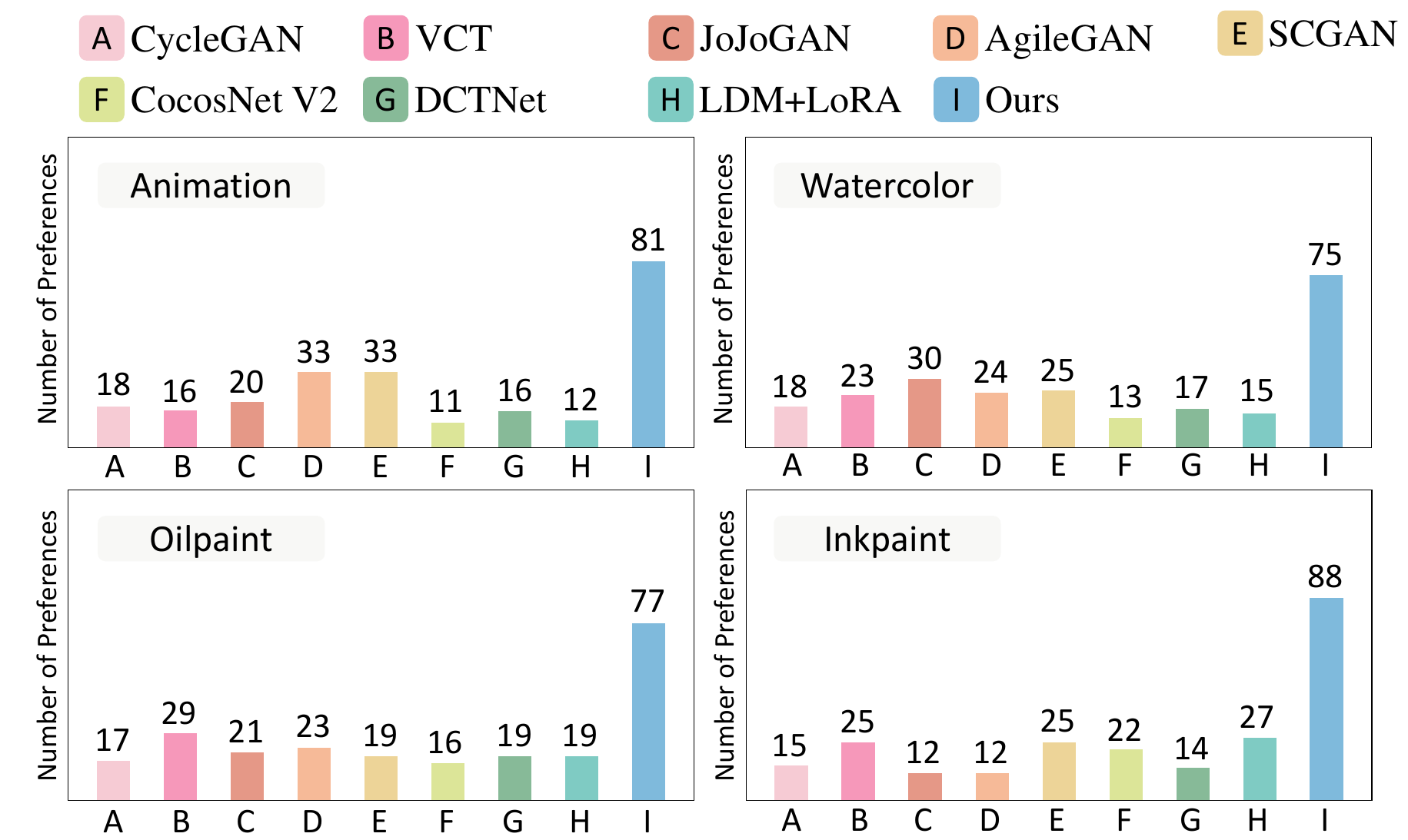}
\vspace{-1.5em}
\caption{Results of user study. Our method is preferred across all styles.}
\label{fig:user_study}
\vspace{-1.5em}
\end{figure}

\subsection{Qualitative Comparison}

We show the comparison of our proposed method and previous methods in Fig.\ref{fig:comparison}, and summarize the strengths and weaknesses of various methods compared to the proposed method:

\textbf{CycleGAN} effectively stylizes portraits for animation and oilpaint style, but can introduce deformations in the  hair area for the animation style and unclear artifacts for the oilpaint style. It also faces challenges, such as mode-collapse, especially noticeable in the watercolor and inkpaint styles due to the limited number of style samples available.

\textbf{AgileGAN} successfully synthesizes visually pleasing results for animation, watercolor and oilpaint. However, it falls short in preserving the identity of the subjects and maintaining background information. Additionally, it struggles to provide meaningful results for inkpaint style.

\textbf{SCGAN} excels in faithfully preserving the identity information of portrait images and effectively transferring style colors. However, it may fall short in transferring sufficient textures to the results.
 
\textbf{DCTNet} produces generally high-quality stylized samples, but some suffer from blurred backgrounds and excessive smoothing. Additionally, the strength of the stylized effects may not always meet expectations.
 
\textbf{CocosNetV2} generates stylized results with the guidance of HED\cite{xie2015holistically} edge map, the loose control hinders the preservation of original identities and even leads to occasional artifacts.
 
For \textbf{LDM}, we use StableDiffusionV1.5 as base model and adopt Low-Rank adaptation training (LoRA) for stylization training, and generate stylization results using ControlNet with HED edge detection. As prompt engineering heavily effects the final results of LDM with LoRA, we use the same prompt "masterpiece, best quality, a boy/girl/man/woman" for all images for fair comparison. The LoRA training leads to stylization results with smooth surface, and the lack of diversity in the style dataset also causes distribution bias. As we are using edge maps detected by HED for spatial control, the colors are not fully preserved and the some details such as haircut and beard are also changed. 

\textbf{JoJoGAN} blurs background due to the StyleGAN inversion \cite{richardson2021encoding} it deploys. Also, it fails to properly reconstruct human faces for inkpaint style as there're only limited number of images in the training set.

\textbf{VCT} tends to cause artifacts on facial characters such as the eyes region of all images in animation style and the little girl in inkpaint style, and the mouth region of the red hair female in oilpaint style.

\textbf{Our proposed method} synthesizes visually pleasant stylization results for all the 4 styles,  because the proposed geometric alignment effectively aligns the key areas and helps the model learn style information of corresponding regions. It also preserves identity, gender, age and even background information, as the geometric aligned cycle-consistent loss effectively constrains the spatial information. To conclude, our method outperforms previous methods in better stylization quality, fewer artifacts, preservation of identity and stability on small training dataset.

We also illustrate the comparison of stylized frames of CycleGAN, SCGAN, DCTNet and LDM with LoRA with the proposed method. As shown in Fig.\ref{fig:frames1}, CycleGAN caused abnormal face shape and exaggerated illumination; The facial region of DCTNet results are darker and details are blurred, and it also fails to stylize the iconic eyes; SCGAN generates results with little stylization effects but high-frequency noises, and the LDM with LoRA causes over smooth textures, strange illumination and the mouth regions are also out of control even if we tried to tune the prompt. Our proposed method, on the contrary, generates properly stylized results with fine details and can be smoothly merged into the background.

\vspace{-0.5em}
\subsection{User Study}
The assessment of portrait stylization is highly subjective and easily influenced by personal preference. To address this issue, we employ user studies to demonstrate how different individuals evaluate our method and previous methods. 30 participants are invited to select the results with the best stylization quality and preserve the person's identity.  We show each participant 32 image sets as part of the evaluation process. These sets were divided into groups of eight, each associated with a specific style. Within each group, participants were asked to choose the result that exhibited the best stylization quality while effectively preserving the identity of the original person. For each image set, participants were provided with the stylized results produced by our method as well as those generated by eight previously established methods. This comprehensive evaluation approach allowed us to collect valuable insights into the user perception of our method's performance across a diverse range of styles and conditions.

The results of user study are visually presented in Fig.\ref{fig:user_study}, where our proposed method received the most numbers of preference among all the methods compared.  To further analyze the comparision, we applied Kruskal-Wallis test as a statistical method. The results of Kruskal-Wallis test clearly demonstrate that our proposed method significantly outperforms all the previous methods in terms of user preference for four specific styles, with a significance level of p \textless 0.05.

\vspace{-0.5em}
\subsection{Ablation Study.}
We show the results of qualitative ablation study in Fig.\ref{fig:component}. Ablating geometric alignment in the generator results in obvious distortions and artifacts in Fig.\ref{fig:component} a, as the generator without TPS lack the representation ability to build the correlation between portrait and style domain, causing the synthesis quality severely deteriorated by the geometric gap. Ablating alignment in discrimination branch causes decreased stylization quality in Fig.\ref{fig:component} b, as the unaligned image pairs bring confusing information for the discriminator. Adopting vanilla feature matching loss leases blurred results in Fig.\ref{fig:component} c, as the lack of spatial information decreases the model's discriminative ability. Ablating auxiliary discriminators of each facial characteristics brings distortions on facial characteristics and artifacts on the skin in \ref{fig:component} d, as the absence of local stylization greatly reduced the diversity of style distribution, and style samples of full faces are not enough to train a high-quality stylization model. In \ref{fig:component} e and f, we show the results of changing cycle loss to MSE loss and DISTS loss \cite{ding2020image}. MSE loss poses less spacial constraints on the images, leading to blurred textures and fewer details. DISTS loss works with similar mechanism as LPIPS loss by calculating multi-scale losses on feature maps extracted by pre-trained CNN, and also achieves similar performance as our full model both qualitatively and quantitatively.

The proposed full framework, which is shown in Fig.\ref{fig:component} e, synthesizes results in apparent animation style with fine textures and clear edges, because the geometrical alignment modules in both generation branch and discrimination branch effectively align the training pairs, enable networks to focus on corresponding regions, avoid distortions caused by misalignments. The spatial-aware feature match loss fully exploits the aligned spatial information, the local stylization increases diversity in the style distribution, and LPIPS works as a powerful spacial constraint, which all facilitate high-quality portrait stylization.

%% file: 5-discussion.tex
 \vspace{-1em}
\section{Limitation and discussion}
In this paper, we observed that building geometric correlations between portraits and style samples can improve the quality of portrait stylization. Based on this finding, we designed a GAN framework that employs geometric alignment for effective training with lightweight generators and limited samples, and introduces spatial aware discrimination and local stylization for improved quality. It outperforms existing methods in qualitative and quantitative assessments, with ablation studies confirming the impact of each component.

Although the proposed method fulfills high-quality portraits stylization  with small datasets and real-time inference on mobile devices, there are still some limitations. For example, severe perspective change of facial images, especially in yaw and pitch, may influence the stylization results. Also, the proposed method is only capable of stylizing the facial region. To address the perspective change issue, incorporating semantic information, such as perspective matching sampling during training, could help improve the robustness. Employing the multi-perspective prior from models trained with massive datasets such as StyleGAN or latent diffusion models may also help. To deal with background stylization, we could employ segmentation models to separate portrait regions and background regions to  separately stylize and merge different regions for final results.